\documentclass[]{ceurart}

\usepackage{booktabs}
\usepackage{multirow}
\usepackage{pifont}
\usepackage{graphicx}
\usepackage[section]{placeins}
\usepackage{tabularx}
\usepackage{xurl}
\usepackage{needspace}
\newcommand{\cmark}{\ding{51}}
\newcommand{\rot}[1]{\rotatebox{90}{\scriptsize\bfseries #1}}
\newcommand{\syskw}[1]{\textit{(#1)}}
\newcommand{\team}[1]{\textit{#1}}
\newcommand{\entryspace}{\Needspace{9\baselineskip}\smallskip}

\begin{document}

\copyrightyear{2026}
\copyrightclause{Copyright for this paper by its authors.
  Use permitted under Creative Commons License Attribution 4.0
  International (CC BY 4.0).}
\conference{CLEF 2026 Working Notes, 21--24 September 2026, Jena, Germany}

\title{Overview of FinMMEval 2026 Task 1: Multilingual Financial Multiple-Choice Question Answering}

\author[1]{Zhuohan Xie}[email=zhuohan.xie@mbzuai.ac.ae]
\cormark[1]
\author[2]{Yuyang Dai}
\author[1]{Rania Elbadry}
\author[3]{Vanshikaa Jani}
\author[4]{Georgi Georgiev}
\author[4]{Dimitar Dimitrov}
\author[6]{Fan Zhang}
\author[5]{Xueqing Peng}
\author[5]{Lingfei Qian}
\author[5]{Jimin Huang}
\author[7]{Jiahui Geng}
\author[1,8]{Yankai Chen}
\author[8,9]{Ye Yuan}
\author[8,9]{Haolun Wu}
\author[2]{Yuxia Wang}
\author[4]{Ivan Koychev}
\author[1]{Veselin Stoyanov}
\author[10]{Mingzi Song}
\author[6]{Yu Chen}
\author[1,8,9]{Xue Liu}
\author[1]{Preslav Nakov}[email=Preslav.Nakov@mbzuai.ac.ae]

\address[1]{Mohamed bin Zayed University of Artificial Intelligence, Abu Dhabi, United Arab Emirates}
\address[2]{INSAIT, Sofia University ``St. Kliment Ohridski'', Sofia, Bulgaria}
\address[3]{University of Arizona, Tucson, United States}
\address[4]{FMI, Sofia University ``St. Kliment Ohridski'', Sofia, Bulgaria}
\address[5]{The Fin AI, United States}
\address[6]{The University of Tokyo, Tokyo, Japan}
\address[7]{Link{\"o}ping University, Link{\"o}ping, Sweden}
\address[8]{McGill University, Montreal, Canada}
\address[9]{Mila, Quebec AI Institute, Montreal, Canada}
\address[10]{Meiji Gakuin University, Tokyo, Japan}
\cortext[1]{Corresponding author.}

\begin{abstract}
FinMMEval 2026 Task 1 evaluates multilingual financial multiple-choice question answering in English, Chinese, Arabic, and Hindi.
The task tests whether systems can select the correct answer to finance questions involving domain terminology, numerical interpretation, and conceptual financial reasoning across languages and scripts.
The final-test set contains 800 questions, with 200 questions per language; gold answers were withheld during submission, and each language was ranked independently by accuracy.
The final leaderboards contain 13 English, 11 Chinese, 11 Arabic, and 10 Hindi ranked submissions.
Top accuracies range from 92.0\% in Hindi to 97.5\% in English and Arabic, with the same leading teams appearing near the top across all four languages.
The documented systems used retrieval augmentation, direct answer-option scoring, language-specific prompting, selective self-consistency, confidence checks, and LLM-based review stages.
\end{abstract}


\maketitle

\section{Introduction}

Financial question answering requires systems to combine domain knowledge with numerical interpretation, reporting conventions, and implicit assumptions.
Depending on the question, a system may need to locate evidence in a company report, perform a calculation, or distinguish between closely related financial concepts.
Multilingual evaluation adds variation in scripts, terminology, and source conventions to these reasoning demands.
Task 1 isolates answer correctness from surface-form variation by requiring systems to return a single option label for each question.
This design evaluates the target capability separately from other properties of generated output~\cite{xie2023nextchapter}.
For each of the four languages---English, Chinese, Arabic, and Hindi---systems answer 200 final-test questions and submit one option for every item.
Providing all systems with the same candidate answers eliminates variation in free-form wording while preserving the need for financial knowledge, numerical calculation, and multilingual terminology.
The task is part of the broader FinMMEval lab at CLEF 2026~\cite{FinMMEval2026}, which also includes short-answer financial QA in Task 2~\cite{FinMMEvalTask2Overview2026} and live trading-agent evaluation in Task 3~\cite{FinMMEvalTask3Overview2026}.

\section{Related Work}

FinMMEval Task 1 is connected to financial QA, finance-oriented language modeling, and multilingual financial evaluation.
FiQA established a retrieval-oriented shared-task setting for financial opinion mining and QA~\cite{maia2018fiqa}, and FinanceBench later framed open-book QA over public-company information as a finance-specific evaluation problem~\cite{islam2023financebench}.
FinQA~\cite{chen2021finqa} and TAT-QA~\cite{zhu2021tatqa} focus on numerical and tabular reasoning over financial reports.
ConvFinQA extends this line to conversational question answering, where systems must track financial evidence across turns~\cite{chen2022convfinqa}.
FinChain emphasizes verifiable reasoning chains in financial tasks~\cite{xie2025finchain}, while RealFin targets implicit assumptions in realistic financial reasoning settings~\cite{dai2026realfin}.
Evidence selection and report construction are also important in finance-oriented QA pipelines.
FinCARDS studies intra-document evidence reranking for financial question answering~\cite{zhou2026fincards}, while FinReporting examines localized financial disclosure reporting~\cite{zhang2026finreporting}.
Herculean extends this broader evaluation direction toward agentic financial intelligence~\cite{peng2026herculean}.
In parallel, finance-specific encoders such as FinBERT~\cite{liu2021finbert} and large language models such as BloombergGPT~\cite{wu2023bloomberggpt} provide modeling foundations for financial NLP.
Instruction-tuned resources such as PIXIU~\cite{xie2023pixiu} and FinGPT~\cite{yang2023fingpt} further broaden the reusable toolset for shared-task systems.
Task 1 also follows the multilingual direction represented by MultiFinBen~\cite{peng2025multifinben}.
SAHM studies Arabic financial-reasoning settings~\cite{elbadry2026sahm}, and BhashaBench provides Hindi/English India-centric domain-evaluation context~\cite{devane2025bhashabenchv1}.

Unlike most free-form financial QA tasks, Task 1 constrains systems to select from predefined options.
This design removes ambiguity in answer normalization and enables exact accuracy scoring while retaining the need for financial knowledge, multilingual terminology, and calculation-oriented reasoning.

\section{Dataset}

\subsection{Data Collection}

The final-test set contains 800 questions, with 200 questions in each of English, Chinese, Arabic, and Hindi.
Each language split consists of normalized financial multiple-choice items with a question, a fixed answer-option set, and a single gold option label.
The selected final-test items and their gold labels were withheld from the FinMMEval development leaderboards, so the portal did not reveal official answers before submission.
The English and Chinese splits were drawn from CFA-style and CPA-style financial exam questions released with RealFin~\cite{dai2026realfin}.
The Arabic split was drawn from the Arabic accounting and finance resources used in SAHM~\cite{elbadry2026sahm}.
The Hindi split was drawn from the Hindi financial multiple-choice resources in BhashaBench~\cite{devane2025bhashabenchv1}.

The four-language design evaluates transfer beyond English-dominant financial QA benchmarks while covering languages with different scripts and resource profiles.
Chinese and Arabic require systems to handle non-Latin scripts and domain terminology; Hindi adds an Indic-language financial QA setting.

\subsection{Data Annotations}

Each released Task 1 item contains a question, a finite set of answer options, and a unique question identifier.
Each item has one gold answer label used for official scoring.
The released data follow a shared multiple-choice schema across the four languages, while final-test labels remained hidden until the official results were computed.
Most final-test items have four answer options; a subset retains two, three, or five options from the source resources instead of being forced into a uniform four-option format.

\subsection{Data Statistics}

The final test contains 200 questions in every language, while the public-development sets range from 70 to 100 items.
Four-choice questions dominate all four splits, ranging from 79.0\% in Chinese to 100\% in Hindi; Arabic is the only split with two-choice items, while English and Chinese retain smaller three- and five-choice subsets (Table~\ref{tab:data}).

\begin{table}[!htbp]
\centering
\caption{Dataset sources, split sizes, and answer-option distributions by language.}
\label{tab:data}
\small
\setlength{\tabcolsep}{3pt}
\begin{tabular}{@{}llrrrrrr@{}}
\toprule
\textbf{Language} & \textbf{Source} & \textbf{Dev} & \textbf{Test} & \textbf{2 opt.} & \textbf{3 opt.} & \textbf{4 opt.} & \textbf{5 opt.} \\
\midrule
English & CFA-style finance exams~\cite{dai2026realfin} & 70 & 200 & 0 & 29 & 159 & 12 \\
Chinese & CPA-style finance exams~\cite{dai2026realfin} & 71 & 200 & 0 & 30 & 158 & 12 \\
Arabic & Arabic finance exams~\cite{elbadry2026sahm} & 97 & 200 & 20 & 0 & 180 & 0 \\
Hindi & Hindi financial MCQs~\cite{devane2025bhashabenchv1} & 100 & 200 & 0 & 0 & 200 & 0 \\
\bottomrule
\end{tabular}
\end{table}

\section{Evaluation Framework}

\subsection{Task Organization}

Task 1 uses a public development release and a hidden final-test release under the same multiple-choice submission schema.
The development leaderboards provided validation scores and baselines, while each final-test submission contained one JSON or JSONL prediction file per language.
Gold answers, scores, and ranks remained hidden until the official results were released.
For a submission to be ranked, its prediction file had to cover the full language split with one valid option label per item; rationales and confidence scores were not evaluated.

\subsection{Evaluation Measures}

The official metric is accuracy:
\[
\mathrm{Accuracy} = \frac{\#\ \mathrm{correct\ answers}}{\#\ \mathrm{test\ questions}}.
\]
Separate language leaderboards preserve language-specific performance and allow teams to participate in a subset of languages.
Only complete, valid 200-item submissions are ranked.
Tied systems retain the published display order but share the same accuracy.
Because the four language splits contain different questions and option-count distributions, their scores are reported independently and should not be interpreted as a controlled comparison of language difficulty.

\subsection{Baselines}

The public-development release included four interpretable baselines: a fixed-seed random selector, an Always-A rule, a round-robin option selector, and a zero-shot Qwen2.5-0.5B-Instruct run.
The random baseline sampled uniformly from the valid answer options shown for each item using a fixed seed.
The Always-A baseline selected option A whenever it was available, and the round-robin baseline cycled through valid option labels in item order.
The Qwen2.5-0.5B-Instruct baseline was run zero-shot without task-specific fine-tuning.
Under uniform random selection, the expected accuracies implied by the option distributions are 25.9\% for English, 26.0\% for Chinese, 27.5\% for Arabic, and 25.0\% for Hindi.
Together, the baselines span chance, simple heuristics, and lightweight zero-shot modeling.

\section{Results and Participant Approaches}

\subsection{Competition Results}

Across the language-specific top-five results in Table~\ref{tab:top-results}, leading accuracies ranged from 92.0\% in Hindi to 97.5\% in English and Arabic.
\team{pjmathematician} ranked first in English, Chinese, and Arabic, while \team{fosu~ltw} and \team{pjmathematician} shared the highest Hindi accuracy.
The same three teams, \team{pjmathematician}, \team{fosu~ltw}, and \team{Pranshu Rastogi}, formed the leading group in every language, although their order and score margins varied.
English attracted the most valid ranked submissions (13), followed by Chinese and Arabic (11 each) and Hindi (10).

\begin{table}[!htbp]
\centering
\caption{Top-five final-test results by language.}
\label{tab:top-results}
\small
\setlength{\tabcolsep}{4pt}
\begin{tabular}{@{}llrrr@{}}
\toprule
\textbf{Language} & \textbf{Team} & \textbf{Order} & \textbf{Correct} & \textbf{Acc. (\%)} \\
\midrule
English & pjmathematician & 1 & 195/200 & 97.5 \\
English & fosu ltw & 2 & 187/200 & 93.5 \\
English & Pranshu Rastogi & 3 & 182/200 & 91.0 \\
English & NLP-DE & 4 & 175/200 & 87.5 \\
English & UyoAngle & 5 & 166/200 & 83.0 \\
\midrule
Chinese & pjmathematician & 1 & 193/200 & 96.5 \\
Chinese & fosu ltw & 2 & 184/200 & 92.0 \\
Chinese & Pranshu Rastogi & 3 & 184/200 & 92.0 \\
Chinese & NLP-DE & 4 & 174/200 & 87.0 \\
Chinese & UyoAngle & 5 & 154/200 & 77.0 \\
\midrule
Arabic & pjmathematician & 1 & 195/200 & 97.5 \\
Arabic & Pranshu Rastogi & 2 & 190/200 & 95.0 \\
Arabic & fosu ltw & 3 & 188/200 & 94.0 \\
Arabic & NLP-DE & 4 & 183/200 & 91.5 \\
Arabic & TCLabs & 5 & 179/200 & 89.5 \\
\midrule
Hindi & fosu ltw & 1 & 184/200 & 92.0 \\
Hindi & pjmathematician & 2 & 184/200 & 92.0 \\
Hindi & Pranshu Rastogi & 3 & 183/200 & 91.5 \\
Hindi & TCLabs & 4 & 180/200 & 90.0 \\
Hindi & UyoAngle & 5 & 174/200 & 87.0 \\
\bottomrule
\end{tabular}
\end{table}

\subsection{Analysis of Results}

The four leaderboards show similar high-end performance but different competitive profiles.
English, Chinese, and Arabic all exceeded 96\% at the top, whereas Hindi topped out at 92.0\%; however, Hindi also had the narrowest top-five spread.
Chinese showed the widest spread, at 19.5 percentage points between first and fifth place, followed by English, Arabic, and Hindi.
These differences describe the submitted systems on four non-parallel question sets; they do not isolate language effects from differences in content or team participation.
\FloatBarrier

\subsection{Participant Approaches}

Table~\ref{tab:participants} summarizes components reported in eight Task 1 Working Notes papers.
Checkmarks indicate components explicitly reported by the authors; blank cells mean that the component was not reported.
Output validation was common, while retrieval, language routing, self-consistency, and ensembling were used more selectively.

\FloatBarrier
\begin{table}[!htbp]
\centering
\caption{Components reported by the eight documented Task 1 systems.}
\label{tab:participants}
\footnotesize
\setlength{\tabcolsep}{2.8pt}
\begin{tabular}{@{}>{\raggedright\arraybackslash}p{0.22\linewidth}|ccc|c|ccc|cc@{}}
\toprule
\multirow{2}{*}[-5.5ex]{\textbf{Team}} & \multicolumn{3}{c|}{\textbf{Backbone}} & \textbf{Evidence} & \multicolumn{3}{c|}{\textbf{Prompting}} & \multicolumn{2}{c}{\textbf{Post-processing}} \\
\cmidrule(lr){2-4}\cmidrule(lr){5-5}\cmidrule(lr){6-8}\cmidrule(l){9-10}
& \rot{Gemini} & \rot{GPT} & \rot{Qwen} & \rot{Search/retrieval} & \rot{Language routing} & \rot{Few-shot} & \rot{Self-consistency} & \rot{Validation} & \rot{Ensemble} \\
\midrule
pjmathematician~\cite{vachharajani2026pjmathematician} & \cmark & & & \cmark & & & & \cmark & \\
fosu ltw~\cite{liang2026fosultw} & & \cmark & & & & & & \cmark & \\
Pranshu Rastogi~\cite{rastogi2026pranshu} & \cmark & & & & \cmark & \cmark & \cmark & \cmark & \\
TCLabs~\cite{pontes2026tclabs} & & \cmark & & \cmark & & & & \cmark & \cmark \\
UyoAngle~\cite{ouyang2026uyoangle} & & \cmark & & & & & & & \\
AI\_TLfanclub~\cite{duc2026aitlfanclub} & & & \cmark & & & & & \cmark & \\
TextSentinels~\cite{durairaj2026textsentinels} & & & \cmark & \cmark & & & & & \cmark \\
DS@GT~\cite{ayela2026dsgt} & & & \cmark & \cmark & \cmark & & & & \\
\bottomrule
\end{tabular}
\end{table}
\FloatBarrier

\entryspace
\noindent\textbf{\team{pjmathematician}~\cite{vachharajani2026pjmathematician}}\\
\syskw{Gemini, Google Search grounding, retrieval augmentation, multiple-choice financial QA}\\
The system uses Gemini with Google Search grounding to provide external context before selecting the final answer option.
Strict answer parsing and lightweight post-processing preserve the exact option-label format required by the evaluation protocol.

\entryspace
\noindent\textbf{\team{fosu ltw}~\cite{liang2026fosultw}}\\
\syskw{GPT-5.5, Doubao Expert Mode, expert review, confidence tiers}\\
The \team{fosu ltw} workflow uses GPT-5.5 to generate initial predictions and Doubao Expert Mode to review financial formulas, accounting standards, policy timelines, and confidence tiers.
The reviewed outputs are converted into single answer labels, while confidence tiers support internal consistency checks.

\entryspace
\noindent\textbf{\team{Pranshu Rastogi}~\cite{rastogi2026pranshu}}\\
\syskw{Gemini, multilingual prompting, few-shot examples, self-consistency}\\
\team{Pranshu Rastogi} uses prompt-engineered Gemini models for Task 1.
The system uses language-specific prompt construction, few-shot examples, confidence-based decision logic, and selective self-consistency.
The prompting strategy is adapted to the language and question type rather than applied uniformly across all four splits.

\entryspace
\noindent\textbf{\team{TCLabs}~\cite{pontes2026tclabs}}\\
\syskw{GPT-4o-mini, multi-agent reasoning, retrieval augmentation, structured prompts}\\
\team{TCLabs} uses GPT-4o-mini within a multi-agent architecture containing specialized financial-reasoning, retrieval, and verification roles.
Structured prompts and debate-style checks screen unsupported answer choices before the final option is returned.

\entryspace
\noindent\textbf{\team{UyoAngle}~\cite{ouyang2026uyoangle}}\\
\syskw{GPT-5.2 zero-shot, Qwen2.5 ablations, tokenizer analysis, confidence routing}\\
The official \team{UyoAngle} submission uses GPT-5.2 in a zero-shot setting with structured single-label output in all four languages.
The paper separately analyzes Qwen2.5-14B-Instruct prompting variants for tokenizer sensitivity, confidence-margin routing, definition retrieval, and parse-failure fallback; these experiments are not components of the official submission.

\entryspace
\noindent\textbf{\team{AI\_TLfanclub}~\cite{duc2026aitlfanclub}}\\
\syskw{Qwen, distillation, answer-label scoring, calibration}\\
\team{AI\_TLfanclub} reports a Task 1 system based on distilled Qwen models for multilingual financial exam-style QA.
The system emphasizes direct answer-label scoring, output calibration, and component selection based on development-set performance.

\Needspace{14\baselineskip}
\entryspace
\noindent\textbf{\team{TextSentinels}~\cite{durairaj2026textsentinels}}\\
\syskw{Qwen3-32B, GPT-OSS-20B, question-type routing, FAISS retrieval}\\
The final system uses Qwen3-32B as both the question-type router and a factual/conceptual scorer, with GPT-OSS-20B as a reasoning-oriented scorer.
It assigns route-dependent FAISS retrieval depth and combines option-wise scores from the two models before returning the selected label.

\entryspace
\noindent\textbf{\team{DS@GT}~\cite{ayela2026dsgt}}\\
\syskw{Qwen3, Qwen2.5, language routing, exemplar retrieval, direct option scoring}\\
\team{DS@GT} routes each language to either Qwen3-14B or Qwen2.5-14B and retrieves solved exemplars from language-specific indices.
Its Retrieval-Augmented Direct Scoring procedure compares next-token probabilities for the available option labels instead of generating an unconstrained answer.
Weighted reciprocal-rank fusion provides a fallback when native-language exemplars are insufficient.

\entryspace
The documented systems pursue the same constrained objective through different combinations of direct option scoring, language-specific prompting or routing, exemplar retrieval, and output control.
Across these papers, model families vary, but several systems separate evidence acquisition or exemplar retrieval from constrained option selection and parsing.

\section{Conclusion and Future Work}

FinMMEval 2026 Task 1 was a shared-task track with a fixed evaluation protocol for financial multiple-choice QA in English, Chinese, Arabic, and Hindi.
The task used hidden gold answers, a fixed answer-label schema, and language-specific accuracy leaderboards.
The top systems achieved high accuracy, but participation, score spread, and top performance differed by language.

Future iterations should report accuracy by question type and financial topic alongside participation by language, helping distinguish language effects from differences in content and the set of participating systems.
A unified multilingual score could complement, rather than replace, language-specific leaderboards if future editions use parallel or otherwise normalized cross-language test designs and attract sufficient four-language participation.

\Needspace{8\baselineskip}
\begin{acknowledgments}
We thank the CLEF 2026 organizers for hosting the lab and supporting the working-notes process.
We also thank all participating teams for submitting systems and providing feedback on the submission portals and evaluation workflow.
We are grateful to Georgi Georgiev for supporting the FinMMEval awards, which helped recognize strong participant submissions across the tasks.
The work of Dimitar Dimitrov and Ivan Koychev is partially supported by the project UNITe BG16RFPR002-1.014-0004 funded by PRIDST and also by the EU NextGenerationEU project, through the National Recovery and Resilience Plan of the Republic of Bulgaria, project SUMMIT, No.~BG-RRP-2.004-0008.
\end{acknowledgments}

\begin{aideclaration}
During the preparation of this work, the authors used OpenAI GPT-5.5 for grammar checks, wording revisions, style improvements, and LaTeX formatting assistance.
\end{aideclaration}

\bibliography{custom}

\end{document}